\documentclass[11pt]{article}

\usepackage[a4paper,margin=1in]{geometry}
\usepackage[T1]{fontenc}
\usepackage[utf8]{inputenc}   
\usepackage{lmodern}          
\usepackage{microtype}

\usepackage{amsmath,amssymb,amsthm,mathtools}

\usepackage{graphicx}         
\usepackage{booktabs}         
\usepackage{array}
\usepackage{caption}
\usepackage{subcaption}

\usepackage{float}

\usepackage{algorithm}
\usepackage{algpseudocode}

\usepackage[numbers,sort&compress]{natbib}

\usepackage[hidelinks]{hyperref}

\newtheorem{theorem}{Theorem}[section]

\theoremstyle{definition}

\newtheorem{remark}[theorem]{remark}

\title{Escaping Local Optima in the Waddington Landscape: A Two-Stage TRPO–PPO Approach for Single-Cell Perturbation Analysis}
\author{Francis Boabang\thanks{Department of Mathematics, Toronto Metropolitan University, Toronto, Canada} \and Samuel Asante Gyamerah\thanks{Department of Mathematics, Toronto Metropolitan University, Toronto, Canada}}
\date{} 

\begin{document}
	\maketitle
	
	\begin{abstract}
		Modeling cellular responses to genetic and chemical perturbations remains a central challenge in single-cell biology. Existing data-driven frameworks have advanced perturbation prediction through variational autoencoders, chemically conditioned autoencoders, and large-scale transformer pretraining. However, most existing models rely exclusively on either in silico perturbation data or experimental perturbation data but rarely integrate both, limiting their ability to generalize and validate predictions across simulated and real biological contexts in a digital twin system. Moreover, the models are prone to local optima in the nonconvex Waddington landscape of cell fate decisions, where poor initialization can trap trajectories in spurious lineages.  In this work, we introduce a two-stage reinforcement learning algorithm for modeling single-cell perturbation. We first compute an explicit natural gradient update using Fisher-vector products and a conjugate gradient solver, scaled by a KL trust-region constraint to provide a safe, curvature-aware first step for the policy. Starting with these preconditioned parameters, we then apply a second phase of proximal policy optimization (PPO) with a KL penalty, exploiting minibatch efficiency to refine the policy.  We demonstrate that this initialization strategy substantially improves generalization on Single-cell RNA sequencing (scRNA-seq) perturbation analysis in a digital twin system.
	\end{abstract}

	\section{Introduction}
	
	Understanding how stem cells make fate decisions is a central question in developmental biology and regenerative medicine. Stem cell differentiation is orchestrated by dense networks of interacting transcription factors (TFs) forming gene regulatory networks (GRNs), which govern the timing and nature of cell state transitions \cite{heydari2022iqcell, Semrau2015}. Insights derived from GRNs during differentiation have enabled more rational design of cell culture systems and have direct implications for cell therapy and regenerative applications \cite{Lipsitz2016, Prochazka2017}. A key example is the use of TFs to reprogram embryonic or adult somatic cells into a pluripotent state, where specific driver TFs can induce a network configuration consistent with pluripotency \cite{Takahashi2006}. These processes can be modelled as executable Boolean GRNs that capture both the network topology and the regulatory rules governing gene interactions, enabling simulation of time-evolving cell states \cite{Dunn2019, Peter2012, Yachie-Kinoshita2018}.
	
	Despite their utility, deriving informative, predictive, and executable GRNs remains challenging. Traditionally, constructing GRNs requires integrating evidence from gene perturbation experiments, a process that is labour-intensive, time-consuming, and costly \cite{Peter2012}. Notable progress has been made through automated formal reasoning, which has successfully identified minimal GRNs underlying naive pluripotency in mice, accurately predicting outcomes for a substantial fraction of new experiments \cite{Dunn2014, Yordanov2016}. However, such approaches have been limited by the scale of data and the reliance on low-throughput measurements.
	
	The introduction of single-cell profiling technologies have made the study of cellular differentiation a lot easier. scRNA-seq provides high-coverage, flexible, and accurate measurements of gene expression, ensuring robust clustering and pseudotime inference \cite{Babtie2017, Fiers2018, Pratapa2020}.  The availability of scRNA-seq datasets ensure that  GRNs are implemented in a more efficient manner. However, there is still a limitation on the use of scRNA-seq. scRNA-seq can offer advantages in data richness and flexibility; however, it is disadvantaged by dropout effects and reduced sensitivity for lowly expressed TFs. Furthermore, the scRNA-seq field lacks an integrated platform for inferring, simulating, and analyzing executable GRNs directly from single-cell transcriptomic data \cite{heydari2022iqcell}.
	
	Machine learning (ML) models, such as deep learning, offer powerful tools for high-dimensional representation learning. ML has been deployed for modeling cellular differentiation, disease progression, and drug response \cite{Palla2025ScalableSG}. These ML models make use of either observational or perturbational data. However, there is limited work that applies ML to gene regulatory network modeling. This can be attributed to nonconvex optimization landscapes, limited interpretability, and difficulties in generalizing to unseen perturbations or novel cell types, often hindering their ability to outperform simpler, more interpretable baselines. 
	
	Reinforcement learning can be used to model gene regulatory networks \cite{fu2025scrl}. Contrary to supervised learning approaches that map inputs to outputs, RL agents learn by actively interacting with an environment, making decisions that maximize long-term rewards. This framework naturally aligns with the challenges of predicting cellular responses to perturbations, where interventions such as gene knockouts, drug treatments, or cytokine stimulations can be viewed as actions applied to a dynamic and high-dimensional cellular state space.
	
\begin{figure}[H]
		\centering
		\includegraphics[width=0.8\textwidth]{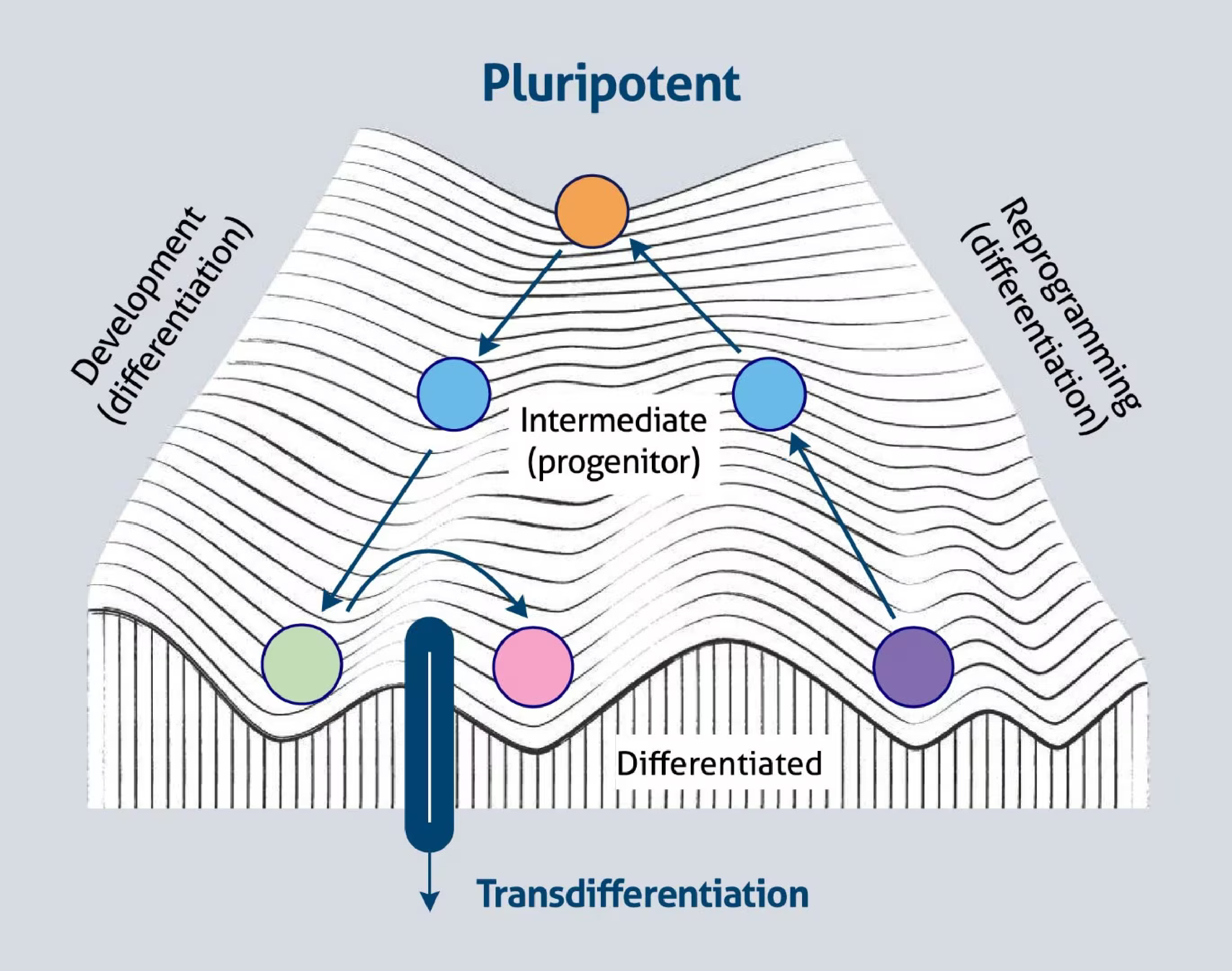}
		\caption{ Waddington’s epigenetic landscapes depict valleys and hills and differentiation, which represent as a ball rolling down branching paths. The shape of the landscape encodes both the stability of cell fates and the potential for plastic transitions between states. The shallow valleys represent high plasticity, which allows cells to respond dynamically to perturbations and environmental cues. Their understanding can help molecular mechanisms reshape the landscape and interpret developmental processes, cellular reprogramming, and disease progression \cite{gilbert1991epigenetic, proteintech2023cell}.}
		\label{fig:landscape}
	\end{figure}

	In the era of single-cell biology, RL  can provide a way for exploring perturbation spaces that are exhaustive and intractable. Instead of combinatorially using brute force to test all possible single and combinatorial perturbations, the RL framework can learn efficient exploration strategies to identify interventions most likely to induce desired cell fate transitions. Next, RL is well-suited for modeling trajectories, serving as a natural tool for approximating the Waddington landscape \cite{gilbert1991epigenetic} and capturing lineage-specific differentiation dynamics as shown in Figure \ref{fig:landscape}.   Again, by framing perturbation analysis as a feedback-driven process, RL facilitates adaptive learning, where agents refine predictions as new single-cell data and perturbation experiments become available.  To further enhance the performance of RL, there is a need to improve the existing policy optimization techniques in reinforcement learning to help the RL model escape local optima and converge to a good lineage or cell fate. 
	
	Cell reprogramming can be interpreted as an optimization trajectory escaping a local optimum under external perturbations such as cytokine, genetic, or environmental.  Genetic or cytokine perturbation can reshape the underlying Waddington landscape of gene regulation. In Waddington's epigenetic landscape, each cell fate represents a local attractor basin stabilized by transcriptional and epigenetic feedback loops. Transitions between these basins require mechanisms that promote exploration and controlled deviation from stability, analogous to optimization strategies in machine learning that prevent premature convergence.
	
	Proximal Policy Optimization in deep reinforcement learning is a first-order optimization algorithm, while Trust Region Policy Optimization (TRPO) \cite{moskovitz2019fop} is second order optimization algorithm. PPO stabilizes policy updates using a clipped (or penalized) surrogate and multiple minibatch passes, giving strong empirical performance with simple implementation, but the proximal behaviour is heuristic, hence it is susceptible to poor generalization performance. TRPO instead enforces an average-KL trust region and computes a natural-gradient step via Fisher-vector products and a conjugate gradient solver, producing principled curvature correction at increased per-update cost. The trust-region updates ensure stable yet effective movement across curvature-sensitive regions of the landscape, entropy regularization promotes diversity and exploration of alternative cell fates, and exploration-driven updates simulate stochastic fluctuations or biological noise that can trigger fate transitions. However, TRPO can be computationally expensive and difficult to scale to large problems due to its second-order nature.  By combining the two techniques in a two-stage fashion, TRPO can provide a warm start to guide PPO in subsequent updates, navigating and reshaping the epigenetic landscape to reach new, biologically meaningful attractor states.
	
	The following existing work has sought to combine the advantages of both approaches. Wang \textit{et al.} propose Trust Region–Guided PPO (TRGPPO), which adapts PPO's clipping range using KL-based trust-region analysis to tighten guarantees while preserving the PPO workflow \cite{wang2019trgppo}. Lascu \textit{et al.} provide a Fisher-Rao geometric reinterpretation of PPO (FR-PPO), deriving geometry-aware surrogates that align PPO updates with Fisher/KL structure; FR-PPO is primarily a theoretical reformulation rather than an explicit natural-gradient solver \cite{lascu2025frppo}. Other lines of work explore first-order preconditioning to cheaply approximate curvature \cite{moskovitz2019fop}, or alter the PPO surrogate to induce greater conservatism (e.g., COPG) \cite{markowitz2023copg}. These works either approximate the curvature or modify the surrogate function, which can lead to a local optimum solution navigating the Waddington landscape.
	
	To address this gap, we introduce a TRPO-preconditioned PPO multistage. This algorithm is different because it is designed sequentially. To be more specific,  we compute a natural-gradient direction using exact Fisher structure (FVP/CG), before scaling it to satisfy a KL budget. Then, it accepts a safe basin of attraction, which is used to initialize a PPO KL penalty and clipped surrogate function. This differs from FR-PPO (no CG/FVP major step), from first-order preconditioning, which utilized the exact Fisher curvature rather than an approximate preconditioner. This initialization strategy ensures that the Fisher-vector product and conjugate gradient step in TRPO computes an approximate natural gradient, making the initial update curvature-aware and aligned with the local geometry of the policy manifold \cite{schulman2015trust}. Moreover, the KL-divergence constraint enforced during the TRPO step ensures a trust-region guarantee, enabling the model to escape local optima and converge to a good solution \cite{schulman2017proximal}. Hence, we fine-tune PPO from a good warm start, ensuring sample efficiency and requiring fewer minibatch updates to achieve performance gains, leading to faster convergence of the model. Lastly,  this initialization design ensures a balance between exploration and stability. Particularly,  the TRPO step preserves policy safety, while the clipped surrogate of PPO refines the policy locally to effectively exploit the available trajectories \cite{schulman2017proximal, lascu2025frppo}. Overall, the two-stage approach leverages the advantages of both TRPO and PPO. 
	
	The contributions of the paper are as follows: 
	
	\begin{enumerate}
		\item We compute an explicit natural gradient step using Fisher-vector products and a conjugate gradient solver, scaled to satisfy a KL trust-region constraint, providing a warm curvature-aware step for the proximal policy optimization.
		\item  We compute preconditioned parameters and proceed to perform the PPO KL penalty fine-tuning phase. Contrary to previous approaches that either approximate curvature or modify the surrogate geometry (e.g., FR-PPO \cite{lascu2025frppo}, first-order preconditioning \cite{moskovitz2019fop}, COPG \cite{markowitz2023copg}), our method explicitly combines exact TRPO preconditioning with PPO's sample-efficient optimization, achieving a tradeoff of both algorithm.
		\item Finally, through extensive evaluation, we demonstrated that this initialization step improves generalization performance in single-cell perturbation analysis.
	\end{enumerate}  
	The remaining parts of the paper are organized as follows. First, we present the related work in Section  \ref{sec:relatedwork}, which is followed by the system model, problem formulation, and methodology in Section \ref{sec:method}. Then, we present the evaluation in Section \ref{sec:evaluation}. We conclude in Section \ref{sec:conclude}.
	
	\section{Related work}
	\label{sec:relatedwork} 
	Machine learning and regulatory networks have improved prediction accuracy in developmental biology, but they can sometimes act as black boxes that do not directly expose causal regulatory logic \cite{scagents2025}.
	
	Tiam et al. focused on the inference and simulation of \emph{executable} gene regulatory networks (GRNs). Their platform (IQCELL) infers logic-based GRNs from pseudo-time ordered scRNA-seq, constructs compact Boolean or logic models constrained by mutual information and temporal ordering, and then simulates developmental trajectories to predict qualitative effects of gene knockouts and perturbations \cite{heydari2022iqcell}. IQCELL emphasizes interpretability and causal hypotheses: the inferred GRNs can recapitulate many experimentally validated causal interactions and enable in-silico perturbation experiments that help study developmental fate decisions. This logic/GRN approach excels at mechanistic insight but can be limited in modelling dose-dependence and continuous dynamics inputs. The drawback of this paper is that since Boolean models define discrete attractor states, cells may converge to an artificial stable region that represents a bad lineage.
	
	Other researchers emphasize data-driven machine learning models that treat perturbation prediction as a conditional mapping from pre-perturbation cell state and perturbation descriptor to post-perturbation expression. Early simple approaches used linear models and tree-based regressors, while more recent methods leverage deep generative models, conditional GANs, and transformer-style architectures to model high-dimensional, heterogeneous single-cell modalities and generalize to unseen perturbations \cite{lotfollahi2019scgen, hetzel2022chemCPA, cui2024scGPT}. A landmark in data-driven approaches is the study of \cite{lotfollahi2019scgen}, who introduced \textit{scGen}, a variational autoencoder framework that predicts how single cells respond to perturbations such as drug treatments or gene knockouts. scGen learns a latent representation of cells and models perturbations as vector arithmetic in latent space, allowing generalization to unseen perturbations and cell types. This work demonstrated that generative neural networks could accurately extrapolate perturbation responses in scRNA-seq, setting the stage for subsequent machine learning approaches.
	
	Also, \cite{lotfollahi2019scgen} proposed \textit{scGen}, a variational autoencoder that learns a latent representation of cells and models perturbations as vector operations in this space. scGen demonstrated that generative neural networks could extrapolate responses of unseen perturbations or cell types, highlighting the power of latent generative modelling for perturbation prediction. However, such VAE-based frameworks implicitly rely on convex optimization heuristics in a highly nonconvex landscape. When viewed through the lens of Waddington’s epigenetic landscape, scGen’s latent arithmetic can become trapped in local optima that correspond to spurious differentiation paths, leading to biologically implausible cell states if initialization is not carefully managed.
	
	To improve generalization, \cite{hetzel2022chemCPA} introduced \textit{chemCPA}, a conditional perturbation autoencoder that integrates molecular structure information with single-cell gene expression data. By conditioning on chemical embeddings and leveraging adversarial training, chemCPA showed improved predictive accuracy for novel compounds and provided mechanistic interpretability for drug responses. Yet, the framework still inherits the same limitations of deep autoencoders: the learned perturbation manifold may become locally biased, forcing cells into rugged valleys of the Waddington landscape. Without mechanisms to escape these local traps, the model risks predicting lineage decisions that correspond to bad minima, cellular fates that do not reflect biological reality.
	
	The latest development in this line is the work of \cite{cui2024scGPT}, who proposed \textit{scGPT}, a foundation model for single-cell multiomics trained at scale with transformer architectures. scGPT offers impressive transferability, enabling cross-modal prediction and perturbation response modelling by pre-training on millions of cells. This represents a conceptual shift toward general-purpose backbones for single-cell biology. However, transformer-based foundation models exacerbate nonconvexity: their immense parameterization can memorize local valleys of the differentiation landscape, and inappropriate initialization or inadequate inductive bias may cause the model to reinforce suboptimal lineage bifurcations. Although scGPT captures global cell state representations, its optimization trajectory does not guarantee exploration of the full Waddington landscape, limiting its ability to faithfully model rare or hard-to-reach differentiation outcomes.
	
	Additionally, \cite{AhlmannEltze2025} reported that deep learning based methods, despite their complexity, did not outperform simple linear baseline models in predicting transcriptome responses to single and double perturbations. The authors attributed this limitation to challenges inherent in nonconvex optimization, where the models failed to converge effectively or differentiate robustly across lineages. This finding highlights the need for more reliable approaches that balance model expressiveness with stability and generalizability. 
	
	In summary, while the above models advance the state of single-cell differentiation modelling through generative latent modelling, chemically informed conditioning, and large-scale pretraining, they all suffer from the fundamental challenge of navigating a rugged, nonconvex Waddington landscape. The risk of becoming trapped in local optima shows as biologically implausible differentiation trajectories, incorrect lineage assignments, or exaggerated drug responses. Overcoming these limitations requires methods that explicitly incorporate nonconvex dynamics and robust initialization strategies to help the perturbation model escape local optima and converge to a good lineage, or cell fate. In addition, most of the existing work relied solely on either experimental perturbation data or in silico perturbation data without combining the two to create a digital twin model.
	
	Recently, a virtual cell challenge was launched \cite{roohani2025virtual}, where participants used different perturbation datasets with available control data (unperturbed data) to build a virtual cell model. Furthermore, they utilize conventional metrics such as differential expression, measure absolute error, and Wasserstein distance to evaluate their model in the virtual cell platform. The participants measured the difference between the predicted perturbation and control for the differential expression score and observed perturbation and predicted perturbation for other regression metrics. However, they failed to model a digital twin of the cell differentiation using both in silico and experimental perturbation datasets to verify if the experimental perturbation was done properly by using metrics such as dynamic time wrapping and Wasserstein distance. By so doing, they can verify if there was a match between in silico perturbation and experimental perturbation. To be more specific, the reproduction of the perturbation experiment using a gene regulatory network-based model was not taken into account in their system model design.    
	
	\begin{figure}[H]
		\centering
		\includegraphics[scale=0.5]{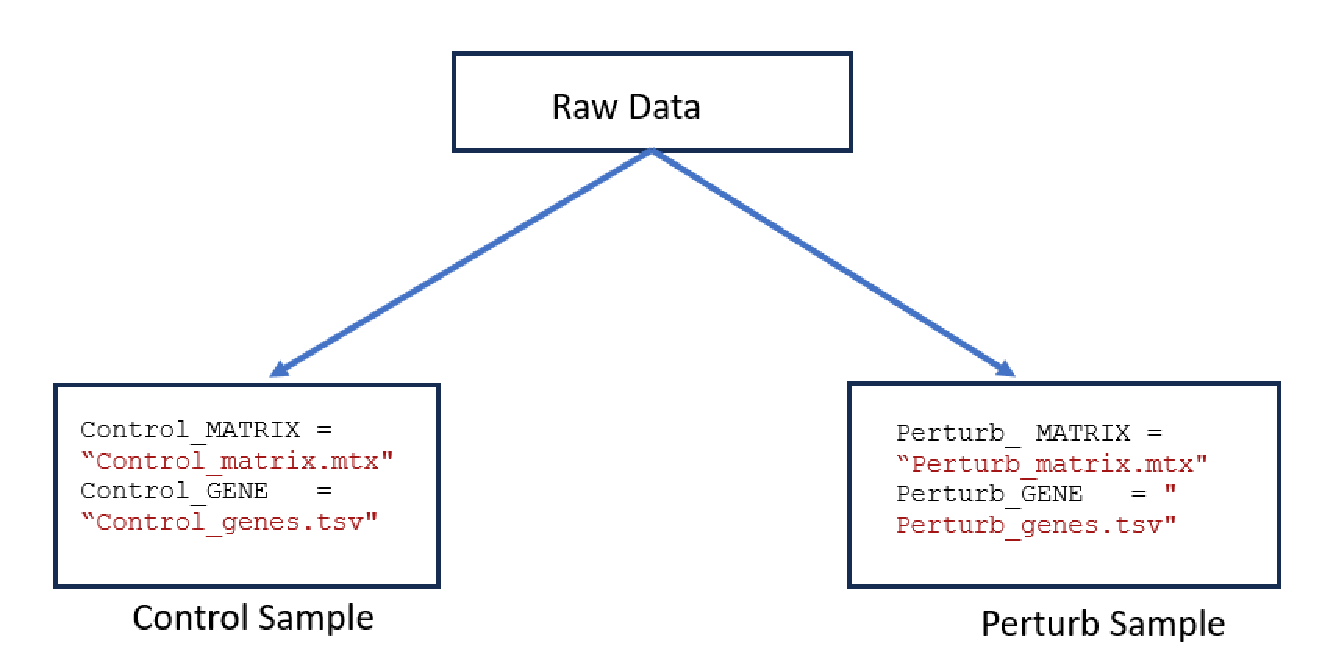}
		\caption{An illustration of a dataset containing the perturbed and unperturbed data \cite{roohani2025virtual}}
		\label{fig:example}
	\end{figure}

	\section{System Model, Problem Formulation, and Methodology}
	\label{sec:method}
	\par
	As input to the proposed digital twin system, we use datasets that contain both the control and perturb data \cite{roohani2025virtual, balmas2025single}. An example of such a dataset is illustrated in Figure \ref{fig:example}. These datasets used various clustering regularly interspaced short palindromic repeats (crispr) screening technologies, such as ShRNA,  crispr-Cas9, crispr knockout, knockdown, and interference to change the epigenetic landscape at different stages of development. For our demonstration, we utilize CRISPR interference screen for the in-silico perturbation modelling.
	
\begin{figure}[H]
		\centering
		\includegraphics[scale=0.4]{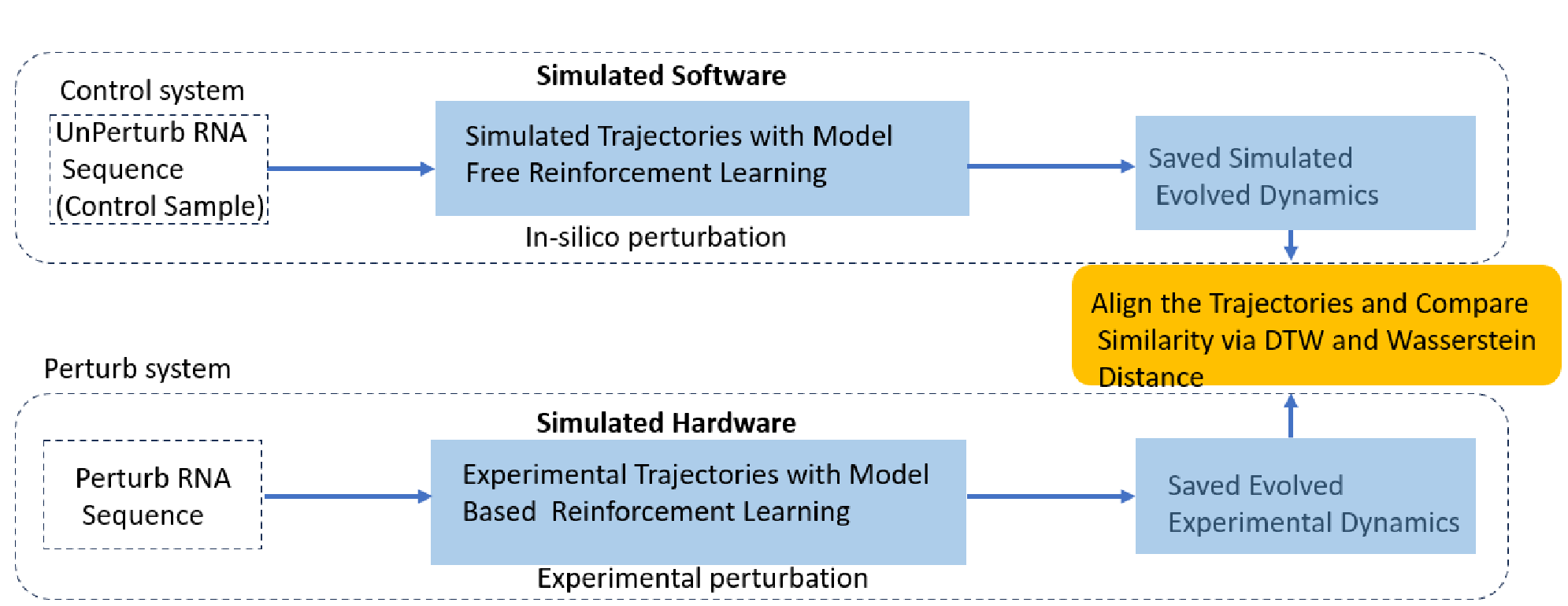}
		\caption{A digital Twin Model of Simulated Software and  Hardware for Single Cell Perturbation Analysis}
		\label{fig:systemmodel}
	\end{figure}

	Our system model, shown in Figure \ref{fig:systemmodel} consist of a hardware model and a software model. The hardware model uses a graph attention neural network model for genes and a perturbation embedding model with no exploration for experimental perturbation learning. The software model consists of the graph attention neural network model for gene embedding. Then,  we used model-free reinforcement learning in a gym environment for in silico perturbation simulation. At the output of the system, we have a dynamic time wrapping and Wasserstein distance to measure the similarities between the saved evolved experimental dynamics trajectories of the hardware system and the saved simulated evolved dynamics trajectories of the simulated software. The pipeline of the proposed digital twin model is described in the link below \footnote{\url{https://colab.research.google.com/drive/1e0zO1X-GQaT0OJ5-DL-6MZQvFfINRBPB?usp=sharing}}. The system model can be customized to target specific gene(s) \cite{gubin2018high, lafleur2019ptpn2} by modifying the model-free reinforcement learning environment of the in-silico perturbation in the pipeline. By so doing, we can perform cytokine perturbation or target specific gene(s), such as ptpn \cite{lafleur2019ptpn2} and anti-PD-1 anti-body or anti-CTLA-4 antibody (aCTLA-4) therapy, or even perform combined therapy \cite{gubin2018high}. Furthermore, the model-free reinforcement learning environment can be modified to perturb specific gene(s) by predefining the gene names.

	\subsection{Embedding Module}
	Let $X \in \mathbb{R}^{n \times d}$ denote the matrix of single-cell profiles, where $n$ is the number of cells and $d$ is the dimensionality of the measured genes. The dataset $\mathcal{D} = \{(x_i, p_i, y_i)\}_{i=1}^N$ and a task description $S$ are given, where $x_i \in \mathbb{R}^d$ represents the pre-perturbation profile of a cell, $p_i \in \mathcal{P}$ denotes the applied perturbation such as genes screen and cytokine perturbations, and $y_i \in \mathbb{R}^{d'}$ corresponds to the observed post-perturbation profile. The dataset is divided into $\mathcal{D}_{\text{train}} = \{(x_i, p_i, y_i)\}_{i=1}^M$ and $\mathcal{D}_{\text{test}} = \{(x_i, p_i, y_i)\}_{i=1}^K$, with $p_i \in \mathcal{P}_{\text{test}}$ and $x_i \in X_{\text{test}}$ denoting held-out perturbations and unseen cell profiles, respectively. The model is denoted as $\theta$. 
	
	For the gene embedding module, we used a graph multihead attention neural network to encode the gene expression. We learn the latent representation of the gene expression profile using an encoder and decoder model. We map the function 
	\[
	f_\theta: \mathbb{R}^d \times \mathcal{P} \rightarrow \mathbb{R}^{d'}
	\]
	parameterized by $\theta$, which predicts the post-perturbation state $y_i$ given an initial cell state $x_i$ and perturbation $p_i$.We use  a learnable encoder 
	\[
	g_\phi: \mathbb{R}^d \rightarrow \mathbb{R}^h
	\]
	with parameters $\phi$, producing latent embeddings $z_i = g_\phi(x_i) \in \mathbb{R}^h$ that preserve the manifold structure of control and perturbed cells, thereby ensuring efficient convergence and accurate prediction of a particular lineage.
	
	\subsection{In-silico Perturbation Modeling Module: TRPO-Preconditioned PPO}
	
	Let $\pi_\theta(a|s)$ denote the policy with parameters $\theta$, and let $\theta_{\text{old}}$ be the parameters used to collect a batch of $T$ on-policy trajectories. Let $\hat A_t$ denote the estimated advantage at time step $t$. Define the importance sampling ratio:
	\begin{equation}
		r_\theta(s_t, a_t) = \frac{\pi_\theta(a_t|s_t)}{\pi_{\theta_{\text{old}}}(a_t|s_t)}.
	\end{equation}
	
	\subsubsection{Surrogate Objective and Gradient}
	The standard policy gradient surrogate is
	\begin{equation}
		L_{\text{surr}}(\theta) = \mathbb{E}_t\big[ r_\theta(s_t, a_t)\, \hat A_t \big].
	\end{equation}
	Its gradient at $\theta_{\text{old}}$ is
	\begin{equation}
		\vartheta = \nabla_\theta L_{\text{surr}}(\theta) \big|_{\theta_{\text{old}}} 
		= \frac{1}{T} \sum_{t=1}^{T} \nabla_\theta \log \pi_\theta(a_t|s_t)\big|_{\theta_{\text{old}}}\, \hat A_t.
	\end{equation}
	
	\subsubsection{Fisher-Vector Product and Conjugate Gradient}
	Define the mean KL divergence:
	\begin{equation}
		\bar{D}_{KL}(\theta_{\text{old}}\Vert \theta) = \frac{1}{T} \sum_{t=1}^{T} D_{KL}\big(\pi_{\theta_{\text{old}}}(\cdot|s_t) \, \Vert \, \pi_\theta(\cdot|s_t) \big),
	\end{equation}
	and its Hessian (Fisher matrix) $H = \nabla^2_\theta \bar{D}_{KL}(\theta_{\text{old}}\Vert \theta) |_{\theta_{\text{old}}}$.  
	We solve $H d = \vartheta$ approximately using matrix-free Conjugate Gradient (CG). For any vector $v$, the Fisher-vector product is
	\begin{equation}
		H v \approx \nabla_\theta \Big( \nabla_\theta \bar{D}_{KL}(\theta) \cdot v \Big) \Big|_{\theta_{\text{old}}} + \lambda_{\text{damp}}\, v,
	\end{equation}
	where $\lambda_{\text{damp}}$ is a small damping constant for numerical stability.
	
	\subsubsection{Natural Gradient Step with KL Scaling}
	The natural step is scaled to satisfy a trust-region KL constraint $\delta$:
	\begin{equation}
		\alpha = \sqrt{\frac{2 \delta}{g^\top d + \epsilon}}, \quad
		\Delta \theta_{\text{nat}} = \alpha d, \quad
		\theta' = \theta_{\text{old}} + \Delta \theta_{\text{nat}},
	\end{equation}
	where $\epsilon$ is a small constant to prevent division by zero. Optionally, a backtracking line search can be applied to ensure $\bar{D}_{KL}(\theta_{\text{old}} \Vert \theta') \le \delta$.
	\subsubsection{PPO KL-Penalty Fine-Tuning}
	
	We start from  the TRPO-updated parameters $\theta'$, then,  we perform a PPO fine-tuning
	using a KL-divergence–regularized surrogate objective.
	The likelihood ratio is defined as
	\begin{equation}
		r_\theta(s_t,a_t)
		=
		\frac{\pi_\theta(a_t \mid s_t)}{\pi_{\theta'}(a_t \mid s_t)} .
	\end{equation}
	
	The KL-penalized PPO objective is
	\begin{equation}
		\begin{split}
			& \mathcal{L}^{\text{KL}}_{\text{PPO}}(\theta)
			=
			- \mathbb{E}_t \big[ r_\theta(s_t,a_t)\, \hat A_t \big] \\ &
			+ \beta\, \mathbb{E}_t \Big[
			D_{\mathrm{KL}}\!\big(
			\pi_{\theta'}(\cdot \mid s_t)
			\,\|\, 
			\pi_\theta(\cdot \mid s_t)
			\big)
			\Big],
		\end{split}
	\end{equation}
	
	where $\beta$ controls the strength of the KL regularization.
	The full fine-tuning objective additionally includes value-function regression
	and entropy regularization:
	\begin{equation}
		\begin{split}
			\mathcal{L}_{\text{PPO}}(\theta)
			=
			&\mathcal{L}^{\text{KL}}_{\text{PPO}}(\theta) \\ &
			+ c_v\, \mathbb{E}_t \big[ \tfrac12 (V_\theta(s_t) - \hat R_t)^2 \big]
			- c_e \, \mathbb{E}_t[\mathcal{H}(\pi_\theta(\cdot \mid s_t))].
		\end{split}
	\end{equation}
	
	Here, $V_\theta(s_t)$ denotes the value-function estimate, $\hat R_t$ the empirical return,
	$\mathcal{H}(\cdot)$ the policy entropy, and $c_v, c_e$ are weighting coefficients.
	Fine-tuning is performed for a small number of minibatch epochs, yielding efficient
	first-order refinement while maintaining proximity to the TRPO trust region,
	as summarized in Algorithm~\ref{alg:trpo_ppo_kl}.
	
	\begin{remark}
		From Algorithm~\ref{alg:trpo_ppo_kl}, PP0 with KL divergence penalty is a soft trust region that does not use the hard KL constraints. Rather, it uses the KL penalty to ensure early stopping when KL  is greater than the target KL. We use the TRPO with the hard KL divergence constraint for a few iteration to obtain an estimate of the KL divergence value, which is used to initialize the PPO with KL divergence penalty to provide a warm-start needed for the model to escape local optima whiles navigating the Waddington landscape.
	\end{remark}
	
	\begin{algorithm}[H]
		\caption{TRPO-Preconditioned PPO with KL-Penalty (Two-stage)}
		\label{alg:trpo_ppo_kl}
		\begin{algorithmic}[1]
			\State Initial policy parameters $\theta_0$, KL trust-region threshold $\delta$, damping coefficient $\lambda_{\text{damp}}$, PPO KL penalty coefficient $\beta$, number of PPO fine-tuning epochs $N_{\text{ppo}}$
			\For{$k = 0,1,2,\dots$}
			\State Collect trajectories $\{(s_t,a_t,r_t)\}_{t=1}^{T}$ using policy $\pi_{\theta_k}$
			\State Compute advantages $\hat{A}_t$ and returns $\hat{R}_t$
			\State Compute policy gradient:
			\[
			\vartheta = \frac{1}{T}\sum_{t=1}^{T}
			\nabla_{\theta} \log \pi_{\theta}(a_t \mid s_t)\,\hat{A}_t
			\Big|_{\theta=\theta_k}
			\]
			\State \textbf{Define} Fisher--Vector Product (FVP):
			\[
			\mathcal{F}(v) =
			\nabla_{\theta}\!\left(
			\nabla_{\theta}\overline{D}_{\mathrm{KL}}(\pi_{\theta_k}\|\pi_{\theta}) \cdot v
			\right)
			+ \lambda_{\text{damp}}\,v
			\]
			\State Solve $\mathcal{F}(d)=\vartheta$ approximately using conjugate gradient
			\State Compute step size:
			\[
			\alpha = \sqrt{\frac{2\delta}{\vartheta^\top d + \varepsilon}}
			\]
			\State TRPO update:
			\[
			\theta' = \theta_k + \alpha d
			\]
			
			\State \textbf{PPO fine-tuning:} initialize $\theta \leftarrow \theta'$
			\For{epoch $=1$ to $N_{\text{ppo}}$}
			\For{each minibatch $\mathcal{B}$}
			\State Compute likelihood ratios:
			\[
			r_t(\theta) =
			\frac{\pi_{\theta}(a_t \mid s_t)}
			{\pi_{\theta'}(a_t \mid s_t)}
			\]
			\State Compute PPO KL-penalty objective:
			\[
			\begin{aligned}
				L^{\mathrm{KL}}(\theta)
				&=
				\mathbb{E}_{t\in\mathcal{B}}
				\left[
				r_t(\theta)\,\hat{A}_t
				\right]
				-
				\beta\,
				\mathbb{E}_{t\in\mathcal{B}}
				\left[
				D_{\mathrm{KL}}\!\left(
				\pi_{\theta'}(\cdot\mid s_t)
				\,\|\, 
				\pi_{\theta}(\cdot\mid s_t)
				\right)
				\right]
			\end{aligned}
			\]
			\State Total loss with value and entropy terms:
			\[
			\begin{aligned}
				L_{\text{total}}
				&=
				-L^{\mathrm{KL}}(\theta)
				+ c_v\,\mathbb{E}\!\left[(V_{\theta}(s_t)-\hat{R}_t)^2\right]
				- c_e\,\mathbb{E}\!\left[H(\pi_{\theta}(\cdot\mid s_t))\right]
			\end{aligned}
			\]
			\State Update $\theta$ using a first-order optimizer
			\EndFor
			\EndFor
			\State Set $\theta_{k+1} = \theta$
			\EndFor
		\end{algorithmic}
	\end{algorithm}

	\subsection{Evaluation Module}
	
	We used an OpenAI Gym environment to simulate the effects of CRISPR interference and cytokine perturbation on individual cells. Each episode represents a single cell, and actions correspond to predicted adjustments in gene expression. The environment incorporates stochastic perturbations, including gene screens or cytokine perturbations, enabling the evaluation of predictive models under realistic, variable gene regulatory dynamics. Rewards are defined to encourage models to accurately approximate observed post-perturbation expression profiles, making this setup suitable for training reinforcement learning agents to predict single-cell responses to genetic interventions.
	
	We train  PPO and a two-stage TRPO\_PPO approach. In the two-stage TRPO\_PPO pipeline, we begin with the warm-starting PPO with parameters learned by TRPO. Training is performed for up to a few timesteps per agent, with reward signals derived from the reduction in mean squared error between the predicted and target expression profiles. For the two-stage TRPO\_PPO method, we used a few steps for the first phase of TRPO phase training and the remaining steps for the PPO phase. 
	
	\subsubsection{Evaluation Metrics Per  System }
	We used the following metrics to measure the prediction performance of the simulated hardware or software model.  RMSE and MSE quantify the square error between the true and predicted gene expression profile. Pearson correlation coefficient (PCC) and $R^2$ quantify the strength between predicted and true expression profiles over all the data. MAE is used to compute the average absolute displacement between the true and predicted gene.
	
	\subsubsection{Similarity Measure}
	We used dynamic time wrapping and Wasserstein distance to measure the similarities between the developmental dynamics trajectories generated by the simulated hardware and software model. DTW can be used to assess the similarities between twotime seriess data with higher accuracy and granularity \cite{forestier2012classification}. The evolved trajectories of both system are aligned and compared using the equations below,
	\begin{equation}
		\begin{split}
			& D[i,j]=\text{cost} \quad + \\
			&  \quad \text{min} \left(D[i-1,j], D[i,j-1], D[i-1,j-1],\right), \\   
		\end{split}
	\end{equation}
	where 
	\begin{equation}
		cost =abs(traj_{sim}[i-1]-traj_{real}[j-1])
	\end{equation}
	and
	$D[i, j]$ represents the minimum cumulative cost to align a trajectory $traj_{sim}[:i]$ with $traj_{real}[:j]$.
	Then, the Wasserstein distance similarity measure can be computed as
	\begin{equation}
		W(traj_{sim}^d, traj_{real}^d)=\frac{1}{T}\sum^T_{i=1}|traj_{sim, i}^d-traj_{real,i}^d|.
	\end{equation}
	
	\section{Evaluation Results}
	\label{sec:evaluation}
	We used diverse datasets for training the systems. Each dataset consists of cell context (cell lines/cell types) for out-of-distribution generalization. Moreover, each dataset has different perturbation types such as gene knockouts, knockdowns, interference, and cytokines to simulate cellular differentiation.  For each system (i.e,. simulated hardware and software system), we split the training and test data into 80\% and 20\%, respectively. For the simulated software, we compare the proposed two-stage algorithm with the PPO baseline for insilico perturbation prediction. We used a total of $150\_000$ steps for training the various algorithms. For TRPO\_PPO, we used $10,000$ steps for training the TRPO and the remaining steps for the PPO phase training. Then,  for the experimental perturbation modeling, we only use the embedding graph attention model to predict the perturbation without exploration with RL\footnote{\url{https://colab.research.google.com/drive/1e0zO1X-GQaT0OJ5-DL-6MZQvFfINRBPB?usp=sharing}}.

	\subsection{Genetic Perturbation}
	
\begin{figure}[H]
		\centering
		\includegraphics[scale=0.5]{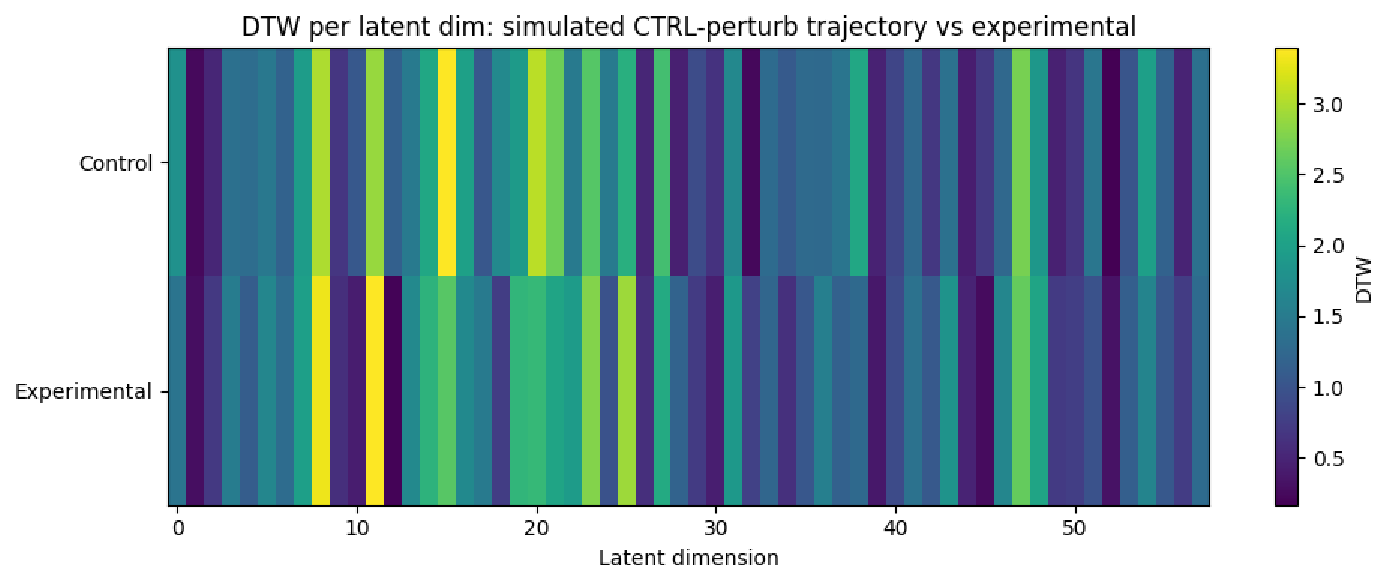}
		\caption{Dynamic time wrapping between the insilico perturbation trajectory and the experimental perturbation trajectory on Replogle genome-wide dataset}
		\label{fig:dtw_replogle}
	\end{figure}

	We showcased the proposed model performance against the baseline on data used for the virtual cell challenge  \cite{roohani2025virtual}. The dataset consists of Replogle genome-wide datasets with the associated perturbations. Furthermore, it contains targeting and non-targeting genes. The non-targeting genes served as the control sample. The dataset is made up of test, train, and validation sets. We extracted a 20k control sample from the training set and a 20k perturbation sample from validation set. We set the top genes to 200. We tested the performance of our proposed TRPO\_PPO against the PPO on the Replogle genome-wide dataset for modeling the insilico perturbation. From Table \ref{ctrl_gene_screen_metrics_replogle}, the result indicated that the proposed TRPO\_PPO greatly outperformed the PPO. It was expected since we used the TRPO to estimate the KL divergence, which was used to initialized the PPO to give it a boost. Also, the results for the embedding layer without the exploration are located in Table \ref{tab:ptpn2_world_model_metrics_replogle_EMBEDDING}. The embedding layer captures the biologically meaningful latent representation of the gene expression. DTW results in Figure \ref{fig:dtw_replogle} as well as mean DTW  and Wasserstein results in Table \ref{tab:trajectory_similarity_mean_replogle} illustrated that the trajectories produced by the insilico and experimental perturbation systems were similar.

	\begin{table}[H]
		\centering
		\caption{Mean regression performance for CTRL gene screening under PPO and two-stage TRPO\_PPO. Metrics are reported for train and test splits on the Replogle genome-wide dataset.}
		\label{ctrl_gene_screen_metrics_replogle}
		\renewcommand{\arraystretch}{1.0}
		\resizebox{0.95\textwidth}{!}{%
			\begin{tabular}{lllrrrrr}
				\hline
				\textbf{System} & \textbf{Algorithm} & \textbf{Split} & \textbf{MSE} & \textbf{RMSE} & \textbf{MAE} & $\mathbf{R^2}$ & \textbf{Pearson} \\
				\hline
				CTRL-Gene-Screen & PPO & Test  & 1.3798 & 1.1551 & 0.9322 & -0.4057 & 0.3379 \\
				CTRL-Gene-Screen & PPO & Train & 1.5284 & 1.2077 & 0.9788 & -0.3784 & 0.3707 \\
				CTRL-Gene-Screen & TRPO\_PPO & Test  & 0.0327 & 0.1678 & 0.1344 & 0.9669 & 0.9906 \\
				CTRL-Gene-Screen & TRPO\_PPO & Train & 0.0361 & 0.1757 & 0.1429 & 0.9672 & 0.9918 \\
				\hline
			\end{tabular}
		}
	\end{table}

	\begin{table}[H]
		\centering
		\caption{Regression performance of the learned world model on the Replogle genome-wide perturbation dataset. Metrics are reported for train and test splits  .}
		\label{tab:ptpn2_world_model_metrics_replogle_EMBEDDING}
		\renewcommand{\arraystretch}{1.0}
		\resizebox{1.0\textwidth}{!}{%
			\begin{tabular}{lllrrrrr}
				\hline
				\textbf{System} & \textbf{Model} & \textbf{Split} & \textbf{MSE} & \textbf{RMSE} & \textbf{MAE} & $\mathbf{R^2}$ & \textbf{Pearson} \\
				\hline
				Experimental Perturbation System & Embedding Model & Test  & 0.1117 & 0.3183 & 0.1969 & 0.8320 & 0.9174 \\
				Experimental Perturbation System & Embedding Model & Train & 0.0528 & 0.2184 & 0.1514 & 0.9326 & 0.9673 \\
				\hline
			\end{tabular}
		}
	\end{table}

	\begin{table}[H]
		\centering
		\caption{Mean trajectory similarity between insilico gene-screen trajectories and experimental references, measured using Dynamic Time Warping (DTW) and Wasserstein distance on the Replogle genome-wide perturbation dataset.}
		\label{tab:trajectory_similarity_mean_replogle}
		\begin{tabular}{lrr}
			\hline
			\textbf{System} & \textbf{DTW} & \textbf{Wasserstein} \\
			\hline
			Insilico Perturbation System  & 1.4028 & 0.6892 \\
			Experimental Perturbation System & 1.3719 & 0.6709 \\
			\hline
		\end{tabular}
	\end{table}
	
\begin{figure}[H]
		\centering
		\includegraphics[scale=0.5]{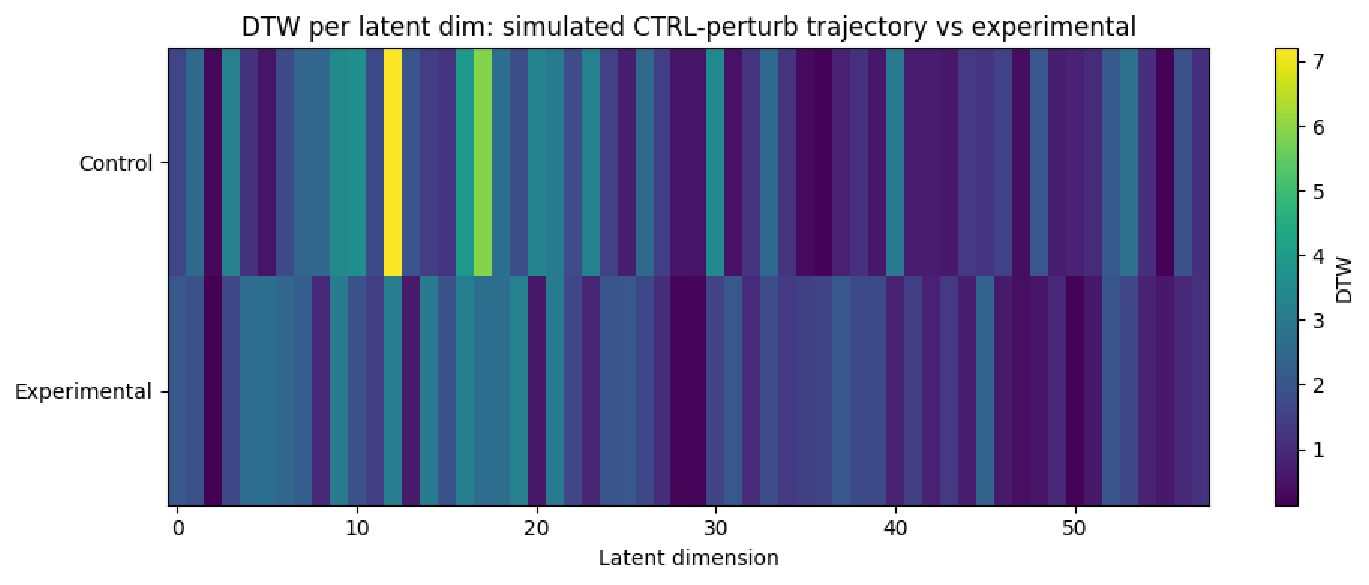}
		\caption{Dynamic time wrapping results comparing insilico perturbation trajectory and experimental perturbation trajectory on Single Cell Transcriptional Perturbome in Pluripotent Stem Cell dataset as the experimental perturbation data and Replogle genome-wide dataset as the control data }
		\label{fig:dtw_smd2}
	\end{figure}

	We leveraged two datasets: a recent dataset on Single Cell Transcriptional Perturbome in Pluripotent Stem Cell Models \cite{balmas2025single} as a perturbation dataset and the Replogle genome-wide dataset as our control dataset for robust evaluation. The Single Cell Transcriptional Perturbome in Pluripotent Stem Cell dataset \footnote{\url{https://zenodo.org/records/15731884}} is made up of unperturbed and perturbed data from  in monolayer cardiomyocytes and organoids samples. We set the top genes very low to 50, to ensure there are common highly variable genes between the control and perturbed data. The results in Table \ref{tab:ctrl_regression_metrics_smad2} showed that TRPO\_PPO outperformed PPO across different metrics. We notice overfitting between the train and test results as reported in Table \ref{tab:ctrl_regression_metrics_smad2}. This can be attributed to the quality of the overlap genes between the very out-of-distribution Replogle genome-wide control and the single-cell transcriptional perturbome in Pluripotent stem cell data. The embedding layer results are presented in Table \ref{model_metrics_smad2_EMBEDDING}. Also, there was a slight variation between the trajectories generated by the insilico and experimental perturbation systems, as illustrated in Figure \ref{fig:dtw_smd2} and Table \ref{tab:trajectory_similarity_mean_smad2}. It was because of the very out-of-distribution control and perturbation sets from different cell contexts and experiments. 
	
	\begin{table}[H]
		\centering
		\caption{Mean regression performance for insilico gene screening on Single Cell Transcriptional Perturbome in the Pluripotent Stem Cell dataset as the experimental perturbation dataset and the Replogle genome-wide dataset as the control data.}
		\label{tab:ctrl_regression_metrics_smad2}
		\begin{tabular}{lllc cccc c}
			\toprule
			System & Algorithm & Split & MSE & RMSE & MAE & $R^2$ & Pearson $r$ \\
			\midrule
			CTRL-Gene-Screen & PPO   & Test  & 0.7984 & 0.8487 & 0.7230 & 0.4116 & 0.8235 \\
			CTRL-Gene-Screen & PPO       & Train & 0.8411 & 0.8670 & 0.7462 & 0.3337 & 0.8384 \\
			CTRL-Gene-Screen & TRPO\_PPO & Test  & 0.1693 & 0.3603 & 0.3033 & 0.8757 & 0.9848 \\
			CTRL-Gene-Screen & TRPO\_PPO & Train & 0.1873 & 0.3830 & 0.3338 & 0.8485 & 0.9857 \\
			\bottomrule
		\end{tabular}
	\end{table}

	\begin{table}[H]
		\centering
		\caption{Perturb Embedding model regression performance on Single Cell Transcriptional Perturbome in the Pluripotent Stem Cell dataset as the experimental perturbation dataset and the Replogle genome-wide dataset as the control data.}
		\label{model_metrics_smad2_EMBEDDING}
		\renewcommand{\arraystretch}{1.0}
		\resizebox{1\textwidth}{!}{%
			\begin{tabular}{lllccccc}
				\toprule
				\textbf{System} & \textbf{Model} & \textbf{Split} & \textbf{MSE} & \textbf{RMSE} & \textbf{MAE} & \textbf{$R^2$} & \textbf{Pearson $r$} \\
				\midrule
				Experimental Perturbation System & Embedding Model & Test  & 0.3733 & 0.5943 & 0.3833 & 0.6954 & 0.8638 \\
				Experimental Perturbation System & Embedding Model & Train & 0.0994 & 0.3045 & 0.2045 & 0.9303 & 0.9657 \\
				\bottomrule
			\end{tabular}%
		}
	\end{table}

	\begin{table}[H]
		\centering
		\caption{Mean trajectory similarity between simulated gene-screen trajectories and experimental data on single-cell transcriptional perturbome in the Pluripotent Stem Cell dataset, with the experimental perturbation dataset as the experimental data and the Replogle genome-wide dataset as the control data.}
		\label{tab:trajectory_similarity_mean_smad2}
		\begin{tabular}{lcc}
			\toprule
			System & DTW & Wasserstein \\
			\midrule
			Insilico Perturbation System  & 1.8152 & 0.7947 \\
			Experimental Perturbation System & 1.5839 & 0.6905 \\
			\bottomrule
		\end{tabular}
	\end{table}

	\subsection{Cytokine Perturbation}
\begin{figure}[H]
		\centering
		\includegraphics[scale=0.5]{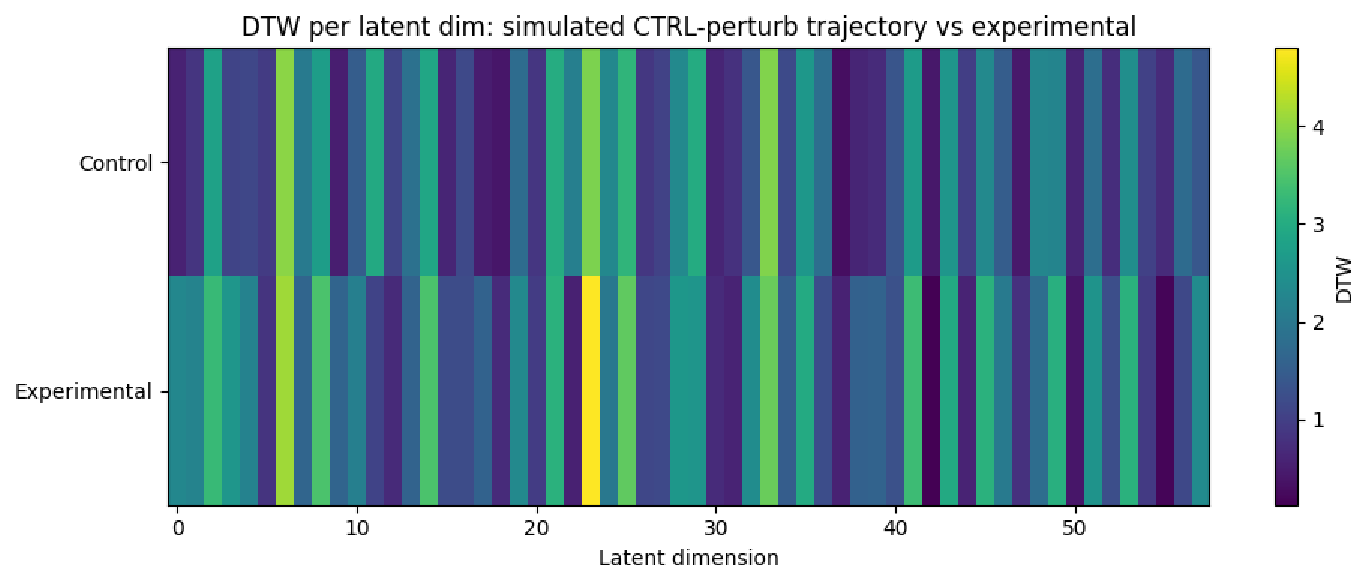}
		\caption{Dynamic time wrapping results comparing in silico perturbation trajectory and Experimental perturbation trajectory on data from parse Biosciences dataset from 12 peripheral Blood Mononuclear Cell donors }
		\label{fig:dtw_cytokine}
	\end{figure}

	We used parse Biosciences dataset from 12 peripheral blood mononuclear cell (PBMC) donors \footnote{\url{https://www.parsebiosciences.com/datasets/10-million-human-pbmcs-in-a-single-experiment/}}. The dataset contains 90 cytokines and phosphate-buffered saline (PBS) control and 18 cell types. We extracted 20k of the subsample from the main dataset. \footnote{\url{https://figshare.com/articles/dataset/pbmc_parse/28589774?file=53372768}.} We set the number of top genes to 100. Table \ref{tab:mean_regression_metrics_cytokine} depicts the mean regression metrics for the insilico perturbation prediction of the various algorithms. We observed that the proposed TRPO\_PPO outperformed the PPO model across all the metrics. We attributed this gain in performance to the initialization strategy provided to PPO, which serve as a warm start to enable the PPO escape local optima and converge to lineage navigating the Waddington landscape. Table \ref{tab:ptpn2_world_model_metrics_cytokine} shows the train and test results of the embedding layer in predicting the experimental perturbation trajectories. The results from the in-silico and experimental perturbation trajectories were compared via DTW and Wasserstein distance as presented in Figure \ref{fig:dtw_cytokine} and Table \ref{tab:trajectory_similarity_mean_cytokine}. The results produced by the two systems were closer to each other.
	
	\begin{table}[H]
		\centering
		\caption{Mean Regression Metrics  of the in-silico cytokine perturbation on control data from parse Biosciences dataset from 12 peripheral Blood Mononuclear Cells donors}
		\label{tab:mean_regression_metrics_cytokine}
		\begin{tabular}{lllccccc}
			\hline
			\textbf{System} & \textbf{Algorithm} & \textbf{Split} & \textbf{MSE} & \textbf{RMSE} & \textbf{MAE} & \textbf{$R^2$} & \textbf{Pearson} \\
			\hline
			CTRL-cytokine & PPO       & Test  & 1.5466 & 1.0380 & 0.9996 & -4.6575 & 0.8389 \\
			CTRL-cytokine & PPO       & Train & 1.5595 & 1.0574 & 1.0045 & -2.4281 & 0.8378 \\
			CTRL-cytokine & TRPO\_PPO & Test  & 0.3918 & 0.5093 & 0.4959 & -0.1304 & 0.9698 \\
			CTRL-cytokine & TRPO\_PPO & Train & 0.3819 & 0.5044 & 0.4882 &  0.0228 & 0.9650 \\
			\hline
		\end{tabular}
	\end{table}

	\begin{table}[H]
		\centering
		\caption{Perturb Embedding Model Regression Metrics   on perturbation dataset from parse Biosciences dataset from 12 peripheral Blood Mononuclear Cells donors }
		\label{tab:ptpn2_world_model_metrics_cytokine}
		\renewcommand{\arraystretch}{1.0}
		\resizebox{1.0\textwidth}{!}{%
			\begin{tabular}{lllccccc}
				\hline
				\textbf{System} & \textbf{Model} & \textbf{Split} & \textbf{MSE} & \textbf{RMSE} & \textbf{MAE} & \textbf{$R^2$} & \textbf{Pearson} \\
				\hline
				Experimental Perturb System &  Embedding Model & Test  & 0.3202 & 0.5525 & 0.3428 & 0.8856 & 0.9479 \\
				Experimental Perturb System & Embedding Model & Train & 0.1030 & 0.3121 & 0.1981 & 0.9667 & 0.9841 \\
				\hline
			\end{tabular}
		}
	\end{table}

	\begin{table}[t]
		\centering
		\caption{Mean Trajectory Similarity Between Simulated and Experimental Trajectories on the  dataset from parse Biosciences dataset from 12 peripheral Blood Mononuclear Cells donors}
		\label{tab:trajectory_similarity_mean_cytokine}
		\begin{tabular}{lcc}
			\hline
			\textbf{System} & \textbf{DTW} & \textbf{Wasserstein} \\
			\hline
			Insilico System  & 1.6428 & 0.4934 \\
			Experimental Perturb System & 1.9082 & 0.6033 \\
			\hline
		\end{tabular}
	\end{table}

	\section{Conclusion}
	\label{sec:conclude}
	We designed a digital twin model for single-cell perturbation analysis. It was made up of simulated hardware and software for measuring the similarities between experimental perturbation trajectories and simulated or insilico perturbation trajectories. Additionally, we introduced a two-stage TRPO\_PPO framework. The two-stage TRPO\_PPO framework provided a general principle for perturbation modeling in systems biology by coupling curvature-aware initialization with efficient local refinement. It was able to enhance convergence in nonconvex settings. Future directions include extending this approach to multi-agent reinforcement learning systems, as well as improving activation functions, optimizers, and loss function formulations to better escape local optima in the Waddington landscape. Also, the proposed framework can be extended beyond transcriptomics to other biological context such as proteomics, multi-omics, and multi-modal. In addition, integrating machine learning methods for the detection of class II antigens could expand the applicability of our framework to immunogenomics and related biomedical domains. Moreover, we can consider a cross-species generalization scenario, where we train on a particular species and test on another species.

	\section*{Acknowledgements}
This research has been supported in part by the Faculty of Science at Toronto Metropolitan University.
	

\begin{thebibliography}{10}

\bibitem{heydari2022iqcell}
Tiam Heydari, Matthew~A. Langley, Cynthia~L. Fisher, Daniel Aguilar-Hidalgo,
  Shreya Shukla, Ayako Yachie-Kinoshita, Michael Hughes, Kelly~M. McNagny, and
  Peter~W. Zandstra.
\newblock Iqcell: A platform for predicting the effect of gene perturbations on
  developmental trajectories using single-cell rna-seq data.
\newblock {\em PLOS Computational Biology}, 18(2):e1009907, 2022.

\bibitem{Semrau2015}
S.~Semrau and A.~van Oudenaarden.
\newblock Studying lineage decision-making in vitro: Emerging concepts and
  novel tools.
\newblock {\em Annu Rev Cell Dev Biol}, 31:317--345, 2015.

\bibitem{Lipsitz2016}
Y.Y. Lipsitz, N.E. Timmins, and P.W. Zandstra.
\newblock Quality cell therapy manufacturing by design.
\newblock {\em Nat Biotechnol}, 34:393--400, 2016.

\bibitem{Prochazka2017}
L.~Prochazka, Y.~Benenson, and P.W. Zandstra.
\newblock Synthetic gene circuits and cellular decision-making in human
  pluripotent stem cells.
\newblock {\em Curr Opin Syst Biol}, 5:93--103, 2017.

\bibitem{Takahashi2006}
K.~Takahashi and S.~Yamanaka.
\newblock Induction of pluripotent stem cells from mouse embryonic and adult
  fibroblast cultures by defined factors.
\newblock {\em Cell}, 126:663--676, 2006.

\bibitem{Dunn2019}
S-J. Dunn, H.~Kugler, and B.~Yordanov.
\newblock Formal analysis of network motifs links structure to function in
  biological programs.
\newblock {\em IEEE/ACM Trans Comput Biol Bioinform}, 2019.

\bibitem{Peter2012}
I.S. Peter, E.~Faure, and E.H. Davidson.
\newblock Predictive computation of genomic logic processing functions in
  embryonic development.
\newblock {\em Proc Natl Acad Sci}, 109:16434--16442, 2012.

\bibitem{Yachie-Kinoshita2018}
A.~Yachie-Kinoshita, K.~Onishi, J.~Ostblom, M.A. Langley, E.~Posfai, and
  J.~Rossant.
\newblock Modeling signaling-dependent pluripotency with boolean logic to
  predict cell fate transitions.
\newblock {\em Mol Syst Biol}, 14, 2018.

\bibitem{Dunn2014}
S-J. Dunn, G.~Martello, B.~Yordanov, S.~Emmott, and A.G. Smith.
\newblock Defining an essential transcription factor program for naive
  pluripotency.
\newblock {\em Science}, 344:1156--1160, 2014.

\bibitem{Yordanov2016}
B.~Yordanov, S-J. Dunn, H.~Kugler, A.~Smith, G.~Martello, and S.~Emmott.
\newblock A method to identify and analyze biological programs through
  automated reasoning.
\newblock {\em Npj Syst Biol Appl}, 2:16010, 2016.

\bibitem{Babtie2017}
A.C. Babtie, T.E. Chan, and M.P.H. Stumpf.
\newblock Learning regulatory models for cell development from single cell
  transcriptomic data.
\newblock {\em Curr Opin Syst Biol}, 5:72--81, 2017.

\bibitem{Fiers2018}
M.W.E.J. Fiers, L.~Minnoye, S.~Aibar, C.~Bravo Gonzalez-Blas, Z.~Kalender~Atak,
  and S.~Aerts.
\newblock Mapping gene regulatory networks from single-cell omics data.
\newblock {\em Brief Funct Genomics}, 17:246--254, 2018.

\bibitem{Pratapa2020}
A.~Pratapa, A.P. Jalihal, J.N. Law, A.~Bharadwaj, and T.M. Murali.
\newblock Benchmarking algorithms for gene regulatory network inference from
  single-cell transcriptomic data.
\newblock {\em Nat Methods}, 2020.

\bibitem{Palla2025ScalableSG}
Giovanni Palla, Sudarshan Babu, Payam Dibaeinia, James~D Pearce, Donghui Li,
  Aly~A. Khan, Theofanis Karaletsos, and Jakub~M. Tomczak.
\newblock Scalable single-cell gene expression generation with latent diffusion
  models.
\newblock {\em ArXiv}, abs/2511.02986, 2025.

\bibitem{fu2025scrl}
Zeyu Fu, Chunlin Chen, Song Wang, Junping Wang, and Shilei Chen.
\newblock scrl: Utilizing reinforcement learning to evaluate fate decisions in
  single-cell data.
\newblock {\em Biology}, 14(6):679, 2025.

\bibitem{gilbert1991epigenetic}
Scott~F Gilbert.
\newblock Epigenetic landscaping: Waddington's use of cell fate bifurcation
  diagrams.
\newblock {\em Biology and Philosophy}, 6(2):135--154, 1991.

\bibitem{moskovitz2019fop}
Ted Moskovitz, Rui Wang, Janice Lan, Sanyam Kapoor, Thomas Miconi, Jason
  Yosinski, and Aditya Rawal.
\newblock First--order preconditioning via hypergradient descent.
\newblock In {\em International Conference on Learning Representations (ICLR)
  -- OpenReview}, 2020.
\newblock ArXiv and OpenReview material (first posted 2019 as
  arXiv:1910.08461).

\bibitem{wang2019trgppo}
Yuhui Wang, Hao He, Xiaoyang Tan, and Yaozhong Gan.
\newblock Trust region--guided proximal policy optimization.
\newblock In {\em Advances in Neural Information Processing Systems (NeurIPS)},
  2019.
\newblock NeurIPS 2019 paper / arXiv preprint.

\bibitem{lascu2025frppo}
Razvan-Andrei Lascu, David {\v S}i{\v s}ka, and {\L}ukasz Szpruch.
\newblock Ppo in the fisher--rao geometry.
\newblock arXiv preprint arXiv:2506.03757, 2025.

\bibitem{markowitz2023copg}
Jared Markowitz and Edward~W. Staley.
\newblock Clipped-objective policy gradients for pessimistic policy
  optimization (copg).
\newblock arXiv preprint arXiv:2311.05846, 2023.

\bibitem{schulman2015trust}
John Schulman, Sergey Levine, Pieter Abbeel, Michael Jordan, and Philipp
  Moritz.
\newblock Trust region policy optimization.
\newblock In {\em Proceedings of the 32nd International Conference on Machine
  Learning (ICML)}, 2015.

\bibitem{schulman2017proximal}
John Schulman, Filip Wolski, Prafulla Dhariwal, Alec Radford, and Oleg Klimov.
\newblock Proximal policy optimization algorithms.
\newblock In {\em arXiv preprint arXiv:1707.06347}, 2017.

\bibitem{proteintech2023cell}
{Proteintech Group}.
\newblock Cell fate commitment and the waddington landscape model.
\newblock
  \url{https://www.ptglab.com/news/blog/cell-fate-commitment-and-the-waddington-landscape-model/},
  2023.
\newblock Accessed: 2025-10-08.

\bibitem{scagents2025}
{Anonymous Authors}.
\newblock scagents: A multi-agent framework for fully autonomous end-to-end
  single-cell perturbation analysis.
\newblock In {\em ICML 2025 Workshop on GenBio (preprint)}, 2025.
\newblock Preprint (ICML submission). Code and models:
  \url{https://anonymous.4open.science/r/scAgents-2025-242E/}.

\bibitem{lotfollahi2019scgen}
M.~Lotfollahi, F.~A. Wolf, and F.~J. Theis.
\newblock scgen predicts single-cell perturbation responses.
\newblock {\em Nature Methods}, 16(8):715--721, 2019.

\bibitem{hetzel2022chemCPA}
Leon Hetzel, Simon Boehm, Niki Kilbertus, Stephan G{\"u}nnemann, Fabian Theis,
  et~al.
\newblock Predicting cellular responses to novel drug perturbations at a
  single-cell resolution.
\newblock {\em Advances in Neural Information Processing Systems},
  35:26711--26722, 2022.

\bibitem{cui2024scGPT}
Haotian Cui, Chloe Wang, Hassaan Maan, Kuan Pang, Fengning Luo, Nan Duan, and
  Bo~Wang.
\newblock scgpt: Toward building a foundation model for single-cell multi-omics
  using generative ai.
\newblock {\em Nature Methods}, 2024.

\bibitem{AhlmannEltze2025}
Constantin Ahlmann-Eltze, Wolfgang Huber, and Simon Anders.
\newblock Deep-learning-based gene perturbation effect prediction does not yet
  outperform simple linear baselines.
\newblock {\em Nature Methods}, 22:1657--1661, 2025.

\bibitem{roohani2025virtual}
Yusuf~H Roohani, Tony~J Hua, Po-Yuan Tung, Lexi~R Bounds, Feiqiao~B Yu,
  Alexander Dobin, Noam Teyssier, Abhinav Adduri, Alden Woodrow, Brian~S
  Plosky, et~al.
\newblock Virtual cell challenge: Toward a turing test for the virtual cell.
\newblock {\em Cell}, 188(13):3370--3374, 2025.

\bibitem{balmas2025single}
Elisa Balmas, Maria~L Ratto, Kirsten~E Snijders, Silvia Becca, Carla Liaci,
  Irene Ricca, Giorgio~R Merlo, Raffaele~A Calogero, Luca Alessandr{\`\i},
  Sasha Mendjan, et~al.
\newblock Single cell transcriptional perturbome in pluripotent stem cell
  models.
\newblock {\em Molecular Systems Biology}, pages 1--49, 2025.

\bibitem{gubin2018high}
Matthew~M Gubin, Ekaterina Esaulova, Jeffrey~P Ward, Olga~N Malkova, Daniele
  Runci, Pamela Wong, Takuro Noguchi, Cora~D Arthur, Wei Meng, Elise Alspach,
  et~al.
\newblock High-dimensional analysis delineates myeloid and lymphoid compartment
  remodeling during successful immune-checkpoint cancer therapy.
\newblock {\em Cell}, 175(4):1014--1030, 2018.

\bibitem{lafleur2019ptpn2}
Martin~W LaFleur, Thao~H Nguyen, Matthew~A Coxe, Brian~C Miller, Kathleen~B
  Yates, Jacob~E Gillis, Debattama~R Sen, Emily~F Gaudiano, Rose Al~Abosy,
  Gordon~J Freeman, et~al.
\newblock Ptpn2 regulates the generation of exhausted cd8+ t cell
  subpopulations and restrains tumor immunity.
\newblock {\em Nature immunology}, 20(10):1335--1347, 2019.

\bibitem{forestier2012classification}
Germain Forestier, Florent Lalys, Laurent Riffaud, Brivael Trelhu, and Pierre
  Jannin.
\newblock Classification of surgical processes using dynamic time warping.
\newblock {\em Journal of biomedical informatics}, 45(2):255--264, 2012.

\end{thebibliography}
	

\end{document}